# Modeling Representation of Videos for Anomaly Detection using Deep Learning: A Review

Yong Shean Chong
Lee Kong Chian Faculty of Engineering and Science
Universiti Tunku Abdul Rahman
yshean@1utar.my

Yong Haur Tay
Lee Kong Chian Faculty of Engineering and Science
Universiti Tunku Abdul Rahman
tayyh@utar.edu.my

### Abstract

*This review article surveys the current progresses made toward video-based anomaly detection. We address the most fundamental aspect for video anomaly detection, that is, video feature representation. Much research works have been done in finding the right representation to perform anomaly detection in video streams accurately with an acceptable false alarm rate. However, this is very challenging due to large variations in environment and human movement, and high space-time complexity due to huge dimensionality of video data. The weakly supervised nature of deep learning algorithms can help in learning representations from the video data itself instead of manually designing the right feature for specific scenes. In this paper, we would like to review the existing methods of modeling video representations using deep learning techniques for the task of anomaly detection and action recognition.*

## 1. Introduction

The ability to detect anomalies in real-time is very valuable, so that appropriate actions can be taken as soon as it is detected to avoid or reduce negative consequences. Thus, many research efforts are done to replace the need of manually detecting anomalous situations, to create an automated video surveillance system. Despite the importance, accurately determining anomalies can be very challenging.

The processing pipeline for such systems usually involves several steps including pre-processing, feature detection and description, sequence or context modeling, and anomaly detection based on certain measure or threshold. Depending on the feature detection method, the pre-processing step might include background subtraction, object detection and tracking. For simplicity, in this paper we only discuss on the feature detection and description methods.

To achieve the objective of automatically detecting anomalous events, some appearance and dynamics of events have to be captured in order to detect the presence of, and identify the spatial location of any anomaly present in the scene. Some examples of conventional feature extractors are optical flow-based descriptors (e.g. HOG3D, HOG/HOF) and trajectory-based descriptors.

Most research works focus on hand engineering features for particular scenes or datasets, but these features need to be manually tuned each time a different scenario is introduced. Meanwhile, deep learning methods are trending in visual-based tasks, due to its capability to produce good representations with raw input. Therefore we put emphasis on reviewing deep learning methods to extract discriminative features from video data.

### 1.1. Challenges in modeling a good representation

High dimensional data such as video cannot be directly fed into a classifier: they contain much redundant information and cause high computational complexity. Therefore video data has to be represented in a way that it can be processed efficiently, yet able to perform accurately on the task given. The key to any successful application is choosing the right representation. However, it is very challenging due to the following reasons:

- *Action pattern variations within the same class*
  The class can be category of the action (e.g. walking, clapping), or classification of the type of event (e.g. normal, abnormal). There exists high diversity of data within one class, due to the variations in style and appearance. The representation should be general to capture the variations in human movements, human-human and human-object interactions.

- *Environmental variations and noise*
  The real-world scenes contain a lot of noise and may vary due to illumination changes and background dynamics. The features must be able to handle the environmental variations for the method to work under noisy environment.

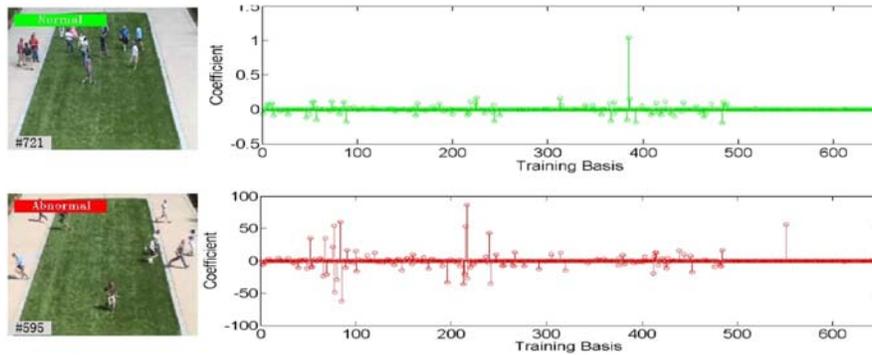

Figure 1. Reconstruction coefficients of normal (top-left) and anomalous (bottom-left) samples, with their corresponding sparse reconstruction coefficients on the right of each sample. Note that the normal sample can be represented as a sparse linear combination of the training bases, while the anomalous sample cannot be reconstructed with the (normal) training bases in a sparse manner. Image adopted from [3].

## 2. Overview of Conventional Features

In this section we will briefly describe some conventional hand-engineered features used for anomaly detection task. We would not discuss the methods in detail since there exists many detailed reviews on these methods [6, 7, 10, 12, 24, 29].

### A. Optical flow-based descriptors

[4] proposed a region-based descriptor called "Motion Context" to describe both motion and appearance information of the spatio-temporal segment. The author uses Edge Orientation Histogram (EOH) as appearance descriptor and Multi-layer Histogram of Optical Flow (MHOF) as motion descriptor. Then for each query spatio-temporal segment, it searches for its best match in the training dataset, and determine the normality using a dynamic threshold. This method is more efficient compared to their previous work using sparse method.

A similar category of descriptors is spatio-temporal video volume descriptors (HOG3D) [11]: these volumes are characterised by the histogram of the spatio-temporal gradient in polar coordinates [23]. In [17], 3D gradient features of each spatio-temporal cube are extracted from the video sequence and trained to obtain sparse combinations with allowable reconstruction errors.

### B. Trajectory based sparse reconstruction

[20] utilises a trajectory based joint sparse reconstruction framework for video anomaly detection, which relies on good tracking to extract trajectories. Inspired by [3] and [4], the authors use Multi-scale HOF (MHOF) as the feature descriptor to construct the basis for sparse representation. A similar approach is proposed by [16]. The fundamental underlying assumption of these methods is that any new feature representation of a normal/anomalous event can be approximately modeled as a (sparse) linear combination of the feature representations of previously observed events in a training dictionary, as visualised by examples shown in Figure 1. For further discussion, the interested reader is directed to an extensive analysis on trajectories [21].

These features require a higher-level modeling in order to determine whether there is an anomaly. Therefore it is difficult to compare and measure the effectiveness of the features chosen --- the experiment results cannot be compared due to different models and measures chosen for anomaly detection. For further discussion on anomaly detection methods such as Dynamic Bayesian Networks and probabilistic topic models, the interested reader is referred to [22].

## 3. Feature Extraction of Video using Deep Learning

The problem of how to represent video sequences is the most fundamental problem in anomaly detection. Instead of introducing the increasingly more complex handcrafted features, recent researches have now moved to using efficient and robust algorithms that learn to extract feature representations from images and videos in a fully unsupervised manner. There are several existing methods for representing images and video sequences using learned features from raw pixel values and frames.

The concept of convolution has been introduced into many existing unsupervised learning algorithms, such as neural networks, Restricted Boltzmann machine (RBM), autoencoders and Independent Component Analysis (ICA). A convolutional architecture comprises training one unsupervised feature extractor on small spatiotemporal patches extracted from sequences of video frames, and subsequently convolving this model with a larger region of the video frames. Eventually, we combine the responses of the convolution step into a single feature vector, which is further processed by a pooling sublayer, to allow for translational invariance. The so-obtained feature vectors may be further presented to a similar subsequent processing layer, thus eventually obtaining a deep training architecture. The stacked model is greedily trained in a layerwise manner, similar to a large number of alternative approaches proposed in the deep learning architecture.

### A. Conditional RBM and Space-Time Deep Belief Network

Conditional RBM (CRBM) [26] is an extension of RBM that models multivariate time-series data. It consists of auto-regressive weights that model short-term temporal structures. The difference between a standard RBM and a CRBM is that in CRBM the hidden units are collected into groups, and each group defines a single weight matrix that is applied to the input image convolutionally to determine the hidden unit activities. The other difference is that a CRBM has a max-pooling step to reduce the spatial res-olution of the hidden layer and to achieve better invariance to spatial transformations of the input image. There are two types of CRBM: spatial CRBM and temporal CRBM.

Space-Time Deep Belief Network (ST-DBN) [2] is a stacked model that uses CRBM as a basic processing unit. The first layer of ST-DBN is a layer of spatial CRBMs, and the second layer is made of temporal CRBMs. ST-DBN can have a multiple stacks of these two layers for further spatial and temporal pooling. Similar to other DBNs, the entire model is trained in a greedy layer-wise manner.

It performs better on discriminative and generative tasks compared to convolutional deep belief networks applied on a per-frame basis. It also has better feature invariance and can integrate information from both space and time to fill in missing data in video.

### B. Independent Component Analysis (ICA) and its variants

A major drawback of existing deep learning architectures for feature extraction concerns the requirement of a priori provision of the number of extracted latent features. This need imposes considerable burden to researchers, as it entails training multiple alternative model configurations to choose from. Methods based on ICA and its variants naturally promote sparsity and allow for automatically inferring the optimal number of latent features.

However, standard ICA [8] has two major drawbacks. First, it is difficult to learn overcomplete feature representations. Second, ICA is sensitive to PCA whitening, which is a common step in reducing the dimensionality of data. As a result it is difficult to scale ICA to high dimensional data.

To overcome the two above-mentioned limitations, reconstruction ICA (RICA) is introduced by [14]. The main difference between RICA and ICA is that in RICA, the hard orthonormal constraint is replaced with a soft reconstruction cost.

According to the author, this change has two benefits: first, it allows unconstrained optimizers to be used instead of relying on slower constrained optimizers (e.g. projected gradient descent) to solve the standard ICA costfunction. Second, the reconstruction penalty works even when the feature representation is overcomplete and the data is not fully white. The mean AP on Hollywood2 dataset shows an improvement of 1.3% when reconstruction penalty is used.

In [1], ICA model is applied to the problem of action recognition in video sequences. Inspired from 3D-CNN [9] and convolutional RBM [28], the author proposes a convolutional model consists of ICA models as its building blocks. The resulted stacked model is similar to other deep learning approaches, except that the hybrid variational inference algorithm for this model is heuristic parameter-free, thus alleviating the need of parameter tuning, which is a major challenge in most deep unsupervised feature extractors.

The author adopts the same pipeline as [5, 11, 13, 30], that is first extract local features (using proposed ICA), then perform vector quantization using K-means, and finally use these feature vectors to train a support vector machine employing a chi-square kernel.

### C. Deeply-Learned Slow Feature Analysis and Gated Models

Slow Feature Analysis (SFA) [25] can learn the invariant and slowly varying features from input signals and has been proved to be valuable in human action recognition [25]. The proposed DL-SFA also adopts max-pooling to capture abstract, structural and translational invariant features. Even though the performance in terms of average precision (AP) is not as good as Hierarchical ISA [15], it is worth noting that without dense sampling the performance of ISA will reduce by about 8% -- 10%.

Another group of models that capture the information of image transformation and correspondence between images [18,19] is gated models. Modeling two images with one set of hidden variables makes it possible to learn spatio-temporal features that can represent the relationship between the images. The key ingredient to make this word is to let the three groups of variables interact multiplicatively, leading to a gated sparse coding model. This gives rise to the gated variants of learning models, such as gated RBM (GRBM), gated autoencoders, etc.

Similar to CRBM, the gated model also makes use of spatial pooling. It learns basis flowfields from each pair of the frames. Result shows that the model has developed sets of local, conditional edge-filters. Given the learned model, it is straightforward to generate dense flowfields [18].

Since GRBM ignores the pictorial structure of images, [28] introduces convolution into GRBM to capture the spatial information. Experiments on human action recognition benchmarks show promising results when compared to handcrafted features.

Table 1. Classification accuracy when using different features on KTH and Hollywood2 dataset.

| Feature used | KTH | Hollywood2 |
| --- | --- | --- |
| HOG3D [11] | 85.3% | 45.3% |
| HOG/HOF [13] | 86.1% | 47.4% |
| HOG/HOF + Mining [5] | **94.5%** | 50.9% |
| Dense trajectories [30] | 94.2% | **58.3%** |
| ST-DBN [2] | 86.6% | - |
| ConvGRBM [28] | 90.0% | 46.6% |
| DL-SFA [25] | 93.1% | 48.1% |
| ISA [15] | 93.9% | 53.3% |
| ICA [1] | **94.3%** | 53.5% |
| RICA [14] | - | **54.6%** |

HOG3D and HOG/HOF are the baseline of conventional methods using hand-engineered features. The current state-of-the-art is HOG/HOF+Mining for KTH dataset and dense trajectories for Hollywood2 dataset. However, note that feature tracking is required for ob-taining dense trajectories. ICA and RICA show promising results in extracting representative and discriminative features for action recognition, even with raw video frames.

## 4. Conclusion and Future Work

We have reviewed the existing approaches and related deep architectures for video representation. Some approaches such as trajectories extraction require identifying and tracking of objects, while optical flow methods do not require such step before features can be extracted from videos. Unsupervised techniques, as opposed to supervised techniques, do not require labeled video data, yet they can be effectively employed for learning good representations.

Despite the largely available features for video domain, we aim to gain useful insights from deep learning models, which have improved performance of many computer vision problems including action recognition. A valuable key insight is that instead of manually design features to produce good results on a particular dataset, deep architectures such as stacked convolutional ISA and RICA can be used to learn good representation for accurate classification.